%% file: main.tex
\def\year{2022}\relax
\documentclass[letterpaper,review,anonymous]{article} 
\usepackage{aaai22}  
\usepackage{times}  
\usepackage{helvet}  
\usepackage{courier}  
\usepackage[hyphens]{url}  
\usepackage{graphicx} 
\usepackage{natbib}  
\usepackage{caption} 
\DeclareCaptionStyle{ruled}{labelfont=normalfont,labelsep=colon,strut=off} 
\usepackage{algorithm}
\usepackage{algorithmic}

\usepackage{newfloat}
\usepackage{listings}
\floatstyle{ruled}
\newfloat{listing}{tb}{lst}{}
\floatname{listing}{Listing}
\usepackage[table]{xcolor}

\usepackage{graphicx}
\usepackage{caption}
\usepackage{subcaption}
\usepackage{pdflscape}
\usepackage{sparklines}
\usepackage{array}

\definecolor{reddit-color}{HTML}{fbd6e3}
\definecolor{twitter-color}{HTML}{e6d4fa}
\definecolor{webmd-color}{HTML}{cff5f5}

\definecolor{my-green}{HTML}{b3e2cd}
\definecolor{my-orange}{HTML}{fdcdac}
\definecolor{my-purple}{HTML}{cbd5e8}
\definecolor{my-pink}{HTML}{f4cae4}
\definecolor{my-yellow}{HTML}{e6f5c9}

\definecolor{my-dark-green}{HTML}{38946b}
\definecolor{my-dark-orange}{HTML}{e05d06}
\definecolor{my-dark-purple}{HTML}{4665a0}
\definecolor{my-dark-pink}{HTML}{be2784}
\definecolor{my-dark-yellow}{HTML}{8dc224}

\usepackage[export]{adjustbox}

\pdfoutput=1

\title{Sensemaking About Contraceptive Methods Across Online Platforms}
\author{
    LeAnn McDowall,\textsuperscript{\rm 1}
    Maria Antoniak,\textsuperscript{\rm 2}
    David Mimno\textsuperscript{\rm 1}
}
\affiliations{
    \textsuperscript{\rm 1}Cornell University \\
    \textsuperscript{\rm 2}Allen Institute for Artificial Intelligence \\

}

\begin{document}

\maketitle

\begin{abstract}
    Selecting a birth control method is a complex healthcare decision.
    While birth control methods provide important benefits, they can also cause unpredictable side effects and be stigmatized, leading many people to seek additional information online, where they can find reviews, advice, hypotheses, and experiences of other birth control users. 
    However, the relationships between their healthcare concerns, sensemaking activities, and online settings are not well understood. 
    We gather texts about birth control shared on Twitter, Reddit, and WebMD---platforms with different affordances, moderation, and audiences---to study where and how birth control is discussed online. 
    Using a combination of topic modeling and hand annotation, we identify and characterize the dominant sensemaking practices across these platforms, and we create lexicons to draw comparisons across birth control methods and side effects.
    We use these to measure variations from survey reports of side effect experiences and method usage.
    Our findings characterize how online platforms are used to make sense of difficult healthcare choices and highlight unmet needs of birth control users.
\end{abstract}

\input{sections/1-introduction}

\input{sections/2-related-work}

\input{sections/3-data}

\input{sections/4-methods}

\input{sections/5-results}

\input{sections/8-discussion}

\input{sections/9-ethics}
\input{sections/10-limitations}

\bibstyle{aaai22}
\bibliography{bibliography}

\end{document}

%% file: sections/1-introduction.tex
\section{Introduction}

Birth control users include nearly two thirds of women in the U.S. between the ages of 15-49---almost 50 million women \citep{daniels2020current}.
In addition to providing a means of family planning, birth control methods can also be used to treat and manage many medical conditions, including acne, endometriosis, cancer risks, gender dysphoria, and irregular and painful menstruation \citep{schindler2013non, nahata2017gender}.
However, birth control methods are not one-size fits all, and the choices of whether to use birth control and which method to select are complicated by personal beliefs, cost and accessibility, and a wide array of side effects that are difficult to predict and identify \citep{manzer2022did,yoost2014understanding,polis2016typical}.

When navigating this decision, birth control users face a sensemaking challenge that involves dealing with social stigma, understanding contraceptive methods and potential side effects, and seeking normalcy after upsetting experiences seeking and using contraception.
This challenge leads many birth control users to the internet \citep{yee2010role,russo2013women}, where they can engage in community sensemaking activities, e.g., seeking and sharing information, reviews, advice, hypotheses, and stories \citep{carron2015help,yang2019seekers}.

Prior work has shown that these various sensemaking activities can vary in frequency across different online platforms \citep{choudhury2014seeking,rivas2020classification,zhang2021distress}.
Different healthcare conditions \citep{sannon2019really} and side effects \citep{choudhury2014seeking} can result in different behavior,
and recent work has demonstrated differences in the online sensemaking practices surrounding reproductive healthcare topics like pregnancy and vulvodynia \citep{young2019girl,andalibi2021sensemaking,chopra2021living}.
However, while birth control researchers have examined the frequencies of different method discussions over time \citep{nobles2019repeal,merz2020population}, a comparison across platforms of the difficult decision-making of birth control users and their sensemaking practices has so far been unexplored.

We study birth control users' unique combination of sensemaking activities---deciding between methods, interpreting side effects, calculating risks, etc.---and we use computational text analysis methods to discover the intersections between these activities and online platforms.
Because different methods and side effects could lead to different activities, we first
explore \textit{which birth control methods and side effects are more likely to be discussed on different online platforms} (\textbf{RQ1}).
Using lexicons crafted for our online settings, we compare these discussion frequencies to the distribution of experienced side effects as reported via a nationally representative survey by \citet{nelson2018women}, indicating increased or decreased interest in discussing these concerns online.
Next, using topic modeling and hand-annotations, we discover and characterize \textit{the kinds of sensemaking activities that birth control users engage in online and how these differ by platform} (\textbf{RQ2}).
We contrast these activities with prior work on related online healthcare communities, and we highlight the particular strategies of birth control users.

\paragraph{Contributions.}
We identify and characterize how birth control users make sense of their experiences and options via three large English-language online platforms (Reddit, Twitter, and WebMD).
Their unique combination of sensemaking strategies included \textit{storytelling}, \textit{risk analysis}, \textit{timing and calculations}, \textit{causal reasoning}, \textit{method and hormone comparison}, and \textit{information and explanations}; in particular, \textit{storytelling} is used to prepare for and overcome painful insertion experiences.
We examine these patterns across birth control methods, side effects, and platforms, unlike prior work, and we compare our observations of side effect discussions to survey data, finding large variations and highlighting the importance of integrating different methods.
Throughout this study, we employ a mixture of lexicons (which we provide as a community resource\footnote{\url{https://github.com/maria-antoniak/birth-control-across-platforms}}), topic modeling, and hand annotation, demonstrating how a detailed and diverse analysis across online platforms can expand our understandings of social sensemaking, web and social media activity, and vital healthcare issues.

%% file: sections/2-related-work.tex
\section{Related Work}
\label{section-related-work}

\smallskip
\noindent\textbf{Sensemaking in online communities.}
As explained by \citet{weick2005organizing}, sensemaking necessarily involves communication and is an ``\textit{activity that talks events and organizations into existence},'' in part through retrospective storytelling.
It is a process that 
relies on collaborative problem solving \citep{pirolli2005sensemaking}, and more recently, it has been defined in closely related work by \citet{andalibi2021sensemaking} 
as ``\textit{how individuals make sense of complex phenomena by constructing mental models that draw on new or existing experiences, information, emotions, ideas, and memories}.''
In online healthcare communities, the individual works to make sense of their healthcare experience, and the community collectively gathers information, compares stories, and makes sense of a shared experience \citep{mamykina2015collective}. 
Sharing experiences can help narrators make sense of their stories \citep{tangherlini2000heroes} and can transfer important information to others without firsthand experience \citep{Bietti2019StorytellingAA}, leading users through a transformative process via the gathering and organizing of information \citep{genuis2017looking,patel2019feel}.
This process is contextual; each community can rely on different strategies \citep{young2019girl}.

\smallskip
\noindent\textbf{Healthcare activity across platforms.}
Specific platform affordances can facilitate different types of healthcare-related disclosure and interactions.
\citet{zhang2021distress} formulate a \textit{social media disclosure ecology} in which platform affordances like anonymity, persistence, and visibility control can predict, e.g., pandemic-related disclosures.
Health information seeking behavior can differ across platforms, with search engines more frequently used for serious and stigmatized conditions and Twitter more frequently used for symptoms \citep{choudhury2014seeking}, and community affordances such as threaded conversations and hashtags can influence participation decisions of those with invisible chronic health conditions \citep{sannon2019really}.
The distribution of content type can also differ by platform; e.g., prior work found that \textit{sharing experiences} is less frequent on social networks than forums, although these rates can depend on healthcare topic \citep{rivas2020classification}.

\smallskip
\noindent\textbf{Reproductive healthcare and online sensemaking.}
Studies of reproductive healthcare communities have emphasized some unique sensemaking practices.
For example, in a study of sensemaking practices after pregnancy loss, \citet{andalibi2021sensemaking} highlighted the importance of seeking emotional validation, rather than just information, when trying to re-discover normalcy.
\citet{young2019girl} similarly supported these dual information management and emotional needs in a study of a vulvodynia Facebook group,
and \citet{chopra2021living} discussed polycystic ovary syndrome and why personal tracking is an important communal activity that involves comparison to others' experiences.
Whether these themes hold for online birth control communities is not currently known.

Two recent studies have focused specifically on online discussions of birth control.
First, \citet{nobles2019repeal} used search queries to track interest in different birth control methods, finding that U.S. political events correlate strongly with increased information seeking behavior online.
Second, \citet{merz2020population} examined the prevalence and sentiment of tweets mentioning different birth control methods. 
The authors found that long-acting methods like the IUD were mentioned more often on Twitter, and the proportion of these tweets increased over time.
While most tweets expressing sentiment about contraception were negative, tweets about long-acting methods were more likely to express positive sentiment.

%% file: sections/3-data.tex
\section{Data}
\label{section:data}

We collected data related to birth control from three prominent online platforms: posts and comments on \textit{r/BirthControl}, birth control reviews on \textit{WebMD}, and \textit{Twitter} posts and replies related to birth control.
These are all English-language subcommunities of larger websites, and while we have limited demographic information, we observe that the majority of location-specific posts (e.g., politics, insurance) are about the U.S.
Table \ref{table:data} summarizes these datasets.

\input{tables/data}

\paragraph{Reddit.}
r/BirthControl\footnote{\url{https://www.reddit.com/r/birthcontrol/}} is a user-created and user-moderated online forum dedicated to birth control.
Users are pseudonymous and range from one-time questioners to experienced question answerers. 
As of November 20, 2022, r/BirthControl had 107,537 members.
According to a Pew Research Center survey of U.S. residents, more men (15\%) than women (8\%) use Reddit, and more White (12\%) and Hispanic (14\%) than Black (4\%) survey respondents use Reddit \citep{perrin2019share},
but these platform-wide distributions are unlikely to be representative of r/BirthControl, for which more detailed demographics are unavailable.
Prior work on a related subreddit found that 81\% of the users identified themselves as white \citep{nobles2020examining}.

Posts tend to be longer than comments, there are more comments than posts in our dataset, and both posts and comments have increased in frequency over time.
For each document, we collected the title (for posts), text, date, and user-assigned tag (for posts).
We removed comments written by the parent post's author, and we also removed stickied comments (auto-generated or mod-written comments).
We did not include documents deleted by the time of collection.

\paragraph{WebMD.}
WebMD\footnote{\url{https://www.webmd.com/sex/birth-control/birth-control-pills}} is a healthcare website that has hosted news, information, prescription discounts, community forums, and user-written medication reviews.
We focused on the medication reviews as a contrast to the Reddit and Twitter datasets.
The majority of reviews are written by women (71\%), and unlike Twitter and Reddit, the website has declined in visitors and reviewers since 2009 \citep{yun2019decline}.
Unlike Twitter and Reddit, WebMD reviews do not allow users to interact directly, though reviews sometimes mention other reviews or the general trends observed in other reviews.

We gathered reviews by querying \emph{birth control} as a condition on the drugs and supplements portion of the site and selecting medications that had more than 5 reviews.\footnote{At the time of data collection, this search term yielded both preventative and emergency contraception, but the terms must be queried separately as of January 2021.}
For each WebMD review, we collect the birth control name, the date of the review, and the review text.

\paragraph{Twitter.}
Twitter is a large social network where discussions range widely from personal to global topics.
Compared to the general public, Twitter users are more likely to be Democrats, and they also skew younger than users of YouTube, Facebook, or Instagram \citep{perrin2019share}.
The gender and racial distribution on Twitter are close to uniform; out of a set of U.S. survey respondents, 24\% of men and 21\% women report using Twitter, while 24\% of Black, 25\% of Hispanic, and  21\% of White respondents report using Twitter \citep{perrin2019share}.
On Twitter, birth control discussions take place in the context of many other discussion topics, and while users can group their conversations using hashtags, there are not separated communities like subreddits.
Moderation is organized by Twitter, rather by users as on Reddit.

Using the Twitter Academic API (v2), we collected all the English Twitter posts containing a set of keywords corresponding to our three target birth control methods.
Separately, we collected all the Twitter replies containing the same set of keywords.
Recent work has shown that this API returns reliable representations of the full tweet space \citep{Pfeffer2022ThisSS}.
See \S Methods below for the design of the Twitter-specific keywords.
We removed Twitter handles from tweet texts.
The high vocabulary size for Twitter posts (Table \ref{table:data}) partly reflects many unique URLs shared in these documents, which can be indicative of information-providing behavior.
There are many possible design decisions when gathering Twitter data, and we chose not to collect the full conversations (replies and parent posts of the keyword-containing posts and replies) to (a) limit the size of our collection, (b) target texts explicitly discussing birth control, and (c) more closely replicate prior work on birth control tweets \citep{merz2020population}.

%% file: tables/data.tex
\begin{table*}[t]

    \centering
    \scriptsize

    \begin{tabular}{p{2cm}p{0.8cm}p{0.8cm}p{0.9cm}p{1.2cm}p{1.5cm}p{1.7cm}p{2.8cm}}

    \hline
    \textbf{Community} & \textbf{\# of \newline Posts} & \textbf{Vocab Size} & \textbf{Mean Tokens} & \textbf{Year Range} & \textbf{Posts Dist. \newline (2007-2020)} & \textbf{Moderation} & \textbf{Structure}  \\
    \hline
    \rule{0pt}{4ex}\colorbox{reddit-color}{Reddit Posts}
    & 68,958
    & 49,088
    & 79
    & 2010-2020
    & 
      \begin{sparkline}{14}
        \sparkspike 0.03333333333333333 0.0
        \sparkspike 0.1 0.0
        \sparkspike 0.16666666666666666 0.0
        \sparkspike 0.23333333333333334 0.0
        \sparkspike 0.3 0.0
        \sparkspike 0.36666666666666664 0.005445744251714401
        \sparkspike 0.43333333333333335 0.042839854780153286
        \sparkspike 0.5 0.07678499394917306
        \sparkspike 0.5666666666666667 0.11308995562726905
        \sparkspike 0.6333333333333333 0.15615671641791046
        \sparkspike 0.7 0.19763513513513514
        \sparkspike 0.7666666666666666 0.3000605082694635
        \sparkspike 0.8333333333333334 0.4996470350947963
        \sparkspike 0.9 0.8377369907220653
        \sparkspike 0.9666666666666667 0.9
      \end{sparkline} 
    & user moderators
    & forum posts
    \\
    \rule{0pt}{4ex}\colorbox{reddit-color}{Reddit Comments}
    & 264,912
    & 67,837
    & 32
    & 2010-2020
    & 
      \begin{sparkline}{14}
        \sparkspike 0.03333333333333333 0.0
        \sparkspike 0.1 0.0
        \sparkspike 0.16666666666666666 0.0
        \sparkspike 0.23333333333333334 0.0
        \sparkspike 0.3 0.0
        \sparkspike 0.36666666666666664 0.00546811055845873
        \sparkspike 0.43333333333333335 0.061820814707548945
        \sparkspike 0.5 0.11436284077315369
        \sparkspike 0.5666666666666667 0.184824970093811
        \sparkspike 0.6333333333333333 0.261208524837877
        \sparkspike 0.7 0.33968866083233645
        \sparkspike 0.7666666666666666 0.48256783982874774
        \sparkspike 0.8333333333333334 0.5731033180129699
        \sparkspike 0.9 0.9
        \sparkspike 0.9666666666666667 0.8297220298432285
      \end{sparkline} 
    & user moderators
    & replies to forum posts
    \\
    \rule{0pt}{4ex}\colorbox{twitter-color}{Twitter Posts}
    & 499,796
    & 398,910
    & 12
    & 2006-2020
    & \begin{sparkline}{14}
        \sparkspike 0.03333333333333333 0.0
        \sparkspike 0.1 0.001111416026344676
        \sparkspike 0.16666666666666666 0.006683932491767288
        \sparkspike 0.23333333333333334 0.09772742864983534
        \sparkspike 0.3 0.38393249176728866
        \sparkspike 0.36666666666666664 0.6351279500548848
        \sparkspike 0.43333333333333335 0.9
        \sparkspike 0.5 0.6975061745334796
        \sparkspike 0.5666666666666667 0.5365669593852909
        \sparkspike 0.6333333333333333 0.7475662047200877
        \sparkspike 0.7 0.7191787870472008
        \sparkspike 0.7666666666666666 0.7062277030735454
        \sparkspike 0.8333333333333334 0.6884759193194292
        \sparkspike 0.9 0.8438271816684961
        \sparkspike 0.9666666666666667 0.7510856888035127
      \end{sparkline} 
    & company
    & tweets (no retweets)
    \\
    \rule{0pt}{4ex}\colorbox{twitter-color}{Twitter Replies}
    & 211,896
    & 73,896
    & 12
    & 2007-2020
    & \begin{sparkline}{14}
        \sparkspike 0.03333333333333333 0.0
        \sparkspike 0.1 0.00010469986040018614
        \sparkspike 0.16666666666666666 0.002006747324336901
        \sparkspike 0.23333333333333334 0.02511051651931131
        \sparkspike 0.3 0.0624709167054444
        \sparkspike 0.36666666666666664 0.1275767798976268
        \sparkspike 0.43333333333333335 0.197987436016752
        \sparkspike 0.5 0.19395649139134483
        \sparkspike 0.5666666666666667 0.20385062819916241
        \sparkspike 0.6333333333333333 0.15432759422987435
        \sparkspike 0.7 0.18976849697533738
        \sparkspike 0.7666666666666666 0.33249185667752446
        \sparkspike 0.8333333333333334 0.4761575151233132
        \sparkspike 0.9 0.8317705909725454
        \sparkspike 0.9666666666666667 0.9
      \end{sparkline} 
    & company
    & replies to tweets
    \\
    \rule{0pt}{4ex}\colorbox{webmd-color}{WebMD Reviews}
    & 18,110
    & 17,487
    & 45
    & 2006-2020
    & \begin{sparkline}{14}
        \sparkspike 0.03333333333333333 0.0
        \sparkspike 0.1 0.1261818181818182
        \sparkspike 0.16666666666666666 0.5632727272727273
        \sparkspike 0.23333333333333334 0.822909090909091
        \sparkspike 0.3 0.762909090909091
        \sparkspike 0.36666666666666664 0.8763636363636365
        \sparkspike 0.43333333333333335 0.9
        \sparkspike 0.5 0.6854545454545455
        \sparkspike 0.5666666666666667 0.592
        \sparkspike 0.6333333333333333 0.39163636363636367
        \sparkspike 0.7 0.34581818181818186
        \sparkspike 0.7666666666666666 0.18
        \sparkspike 0.8333333333333334 0.14363636363636365
        \sparkspike 0.9 0.12654545454545457
        \sparkspike 0.9666666666666667 0.06872727272727273
    \end{sparkline} 
    & company
    & reviews (no reply structure)
    \\[2ex]
    \hline
    \end{tabular}
    \caption{Overview of the three datasets used for comparison, including only texts mentioning our target birth control methods.}
    \label{table:data}
\end{table*}

%% file: sections/4-methods.tex
\section{Methods}

\subsection{Birth Control Method Lexicon}
\label{section:method-labeling}

To explore the prevalence across platforms of different birth control methods, we develop a lexicon to match each document to the primary method being discussed.
Due to data scarcity and space limitations, we do not consider methods that are not available on all three platforms (e.g., WebMD does not include the male condom), and of the remaining methods, we limit our analysis to the three most commonly discussed reversible, non-emergency methods, mirroring prior work such as \citet{merz2020population}.\footnote{Using the lexicons described above for Reddit, we find that the implant (8,578 posts) is discussed more than twice as often as the shot (3,529 posts) or barrier methods like the male condom (3,029 posts). This contrasts with reported rates of contraception use in the U.S., where the pill is used by 14.0\%, the IUD by 8.4\%, the male condom by 8.4\%, the implant by 2.0\%, and the shot by 2.0\% of women aged 15-49 \citep{daniels2020current}.}
Our goal is to measure \textit{prevalence of discussions} rather than \textit{prevalence of usage} because (1) we are interested in the level of interest and concern of birth control users regardless of actual usage and (2) other study designs (e.g., surveys) are better suited to studying usage rates.

We rely on sets of keywords to assign each text a primary birth control method.
Because of their different structures, each platform requires its own set of keywords and matching techniques.
All of these lexicons are made available for future research.
To estimate the lexicon performance, we checked 450 documents balanced across Reddit posts, Twitter posts, and Twitter replies.\footnote{We omitted Reddit comments because of our method detection strategy, which defaulted to the method in the parent post if no method was found in the comment; and we omitted WebMD reviews because our method detection strategy already used hand-labeling of medication tags.}
Our precision and recall scores were perfect (1.0).

\subsubsection{WebMD.}
The URLs for WebMD reviews include a \emph{drugname} parameter.
This parameter includes the medication name followed by its class. 
We were able to classify 99.2\% of the WebMD reviews by mapping these names to their corresponding method, and the remaining reviews were for methods outside our target three.

\subsubsection{Reddit.}
\label{subsubsection:reddit-method-labeling}
To assign Reddit posts and comments to the birth control methods, we use a custom set of keywords.
In all cases, the text was assigned to the method that had the highest keyword count (of all the methods).\footnote{We could instead have assigned each Reddit post and comment to as many methods as were mentioned in its text, rather than assigning each post and comment only to the method mentioned most frequently. 
We chose the latter assignment because it better aligns with the WebMD assignments; WebMD reviews are explicitly assigned by the user to a single method (even if other methods are also mentioned).
We do not find that these different assignment techniques affect our distributional results.}
If the highest keyword count was not for one of our target three methods, or if multiple methods were mentioned with equal frequency, we discarded the text.
We follow a similar procedure for Reddit comments, but in cases where the comments do not contain a birth control keyword, we assign the comment to the birth control method of its parent post.

We developed the Reddit methods keyword set by (a) drawing terms from the WebMD keyword set, (b) drawing terms from the Twitter keyword set,
(c) supplementing the WebMD and Twitter sets with general terms that are overloaded on Twitter but usable on Reddit (e.g., ``pill''), and (d) iteratively running the assignments and examining the unassigned posts and comments.
We were able to assign all but 4.6\% of the Reddit posts and 32.6\% of the Reddit comments to a birth control method.
Manual examination of the unassigned posts reveals discussions of pregnancy scares, access (e.g., online appointments and prescriptions), and treatment of side effects with non-contraceptive medications.
The remainder of these unassigned posts discussed topics that are adjacent to birth control, e.g., insurance, menstruation, but are not explicitly connected to a birth control method.

\subsubsection{Twitter.}
We use a more limited keyword set, derived from a similar survey of birth control tweets by \citet{merz2020population}, to query for tweets about each of our target birth control methods.
Our focus was on precision rather than recall, as the Twitter API requires a keyword match to retrieve tweets on a particular topic; if we included keywords like ``pill'', our results would contain many false positives, unlike Reddit and WebMD where the topic is already constrained to birth control.
After gathering our initial Twitter dataset, we then apply the full Reddit keyword set, using the same methods described for Reddit above.
As with the Reddit data, we find that more texts can be assigned to only the pill, IUD, or implant than to a combination of methods; 12.7\% of posts and 17.6\% of the gathered tweets either mentioned multiple methods with equal frequency or most frequently mentioned a method not in our set of three target methods.

\subsection{Side Effects Lexicon}
\label{section:side-effects}

To measure the frequency of discussions about side effect birth control, we develop a lexicon of terms and patterns.
Because we do not use the side effects lexicon for data collection (unlike the methods lexicon), we can rely on one lexicon across our three datasets.
As with the methods lexicon, we focus specifically on measuring the \textit{prevalence of discussions} rather than the \textit{prevalence of experiences}, as these are better measured via surveys \citep{nelson2018women}.
The frequency of discussions could be influenced by the prevalence of experiences and the level of \textit{concern} birth control users on this platforms have about particular side effects, as well as platform setting and a user's individual goals.
We select patterns that match any discussion of the side effect, whether or not it is mentioned in the affirmative.

We grounded our development of the side effects lexicon in prior work.
We matched the side effect categories from \citet{nelson2018women} as closely as possible; this study conducted a nationally representative survey of U.S. birth control experiences and reported prevalences of side effect experiences across different birth control methods.
In addition to these, we added lexicon categories for pain, skin changes, PMS, appetite changes, sexual partner feeling IUD strings, and heart attack.
We identified these additional categories and patterns from topic models trained on our datasets and from other work; for example, we also drew on work by \citet{barr2010managing} that discusses a variety of side effects and their known frequencies and associations with different birth control methods.
For each side effect, we then iteratively queried and made updates to the lexicon when we encountered false positives or false negatives.

\input{tables/lexicon_coverage}

\subsubsection{Lexicon evaluation.}

To evaluate our lexicon, for each side effect we randomly selected a set of matching and non-matching texts, balanced across platforms and match status, resulting in a set of 1,040 texts.
We then manually checked whether each text does or does not contain a discussion of the target side effect.
Across the side effects, we found precision of 0.98 and recall of 0.96.
We show the lexicon coverage in Table \ref{table:lexicon-coverage}.

\subsubsection{Comparing observations with survey responses.}

We compare the distributions of side effect mentions to prior work surveying people in the U.S. about their experiences with birth control side effects.
This allows us to compare the frequency of side effect experiences with the frequency of online discussions.
Differences between these distributions can indicate healthcare needs that users are addressing via the internet.
We compare to surveys (rather than controlled studies) because surveys better approximate the self-reported anecdotes and personal experiences shared online.
We first converted each distribution to a ranking, where lower ranks represent greater percents of discussions $r_{platforms}$ or experiences $r_{survey}$.
We then find the difference between the ranks for each side effect, $d_{ranks}$.
\begin{equation}
    d_{ranks} =  r_{survey} - r_{platforms}
\end{equation}
These ranks avoid issues in directly comparing percents, which might on average be higher or lower.
When $d_{ranks}$ is positive, it indicates more online discussion than expected given the frequency of reported side effect experiences.

\begin{figure*}[t]
    \centering
    \includegraphics[width=0.49\textwidth]{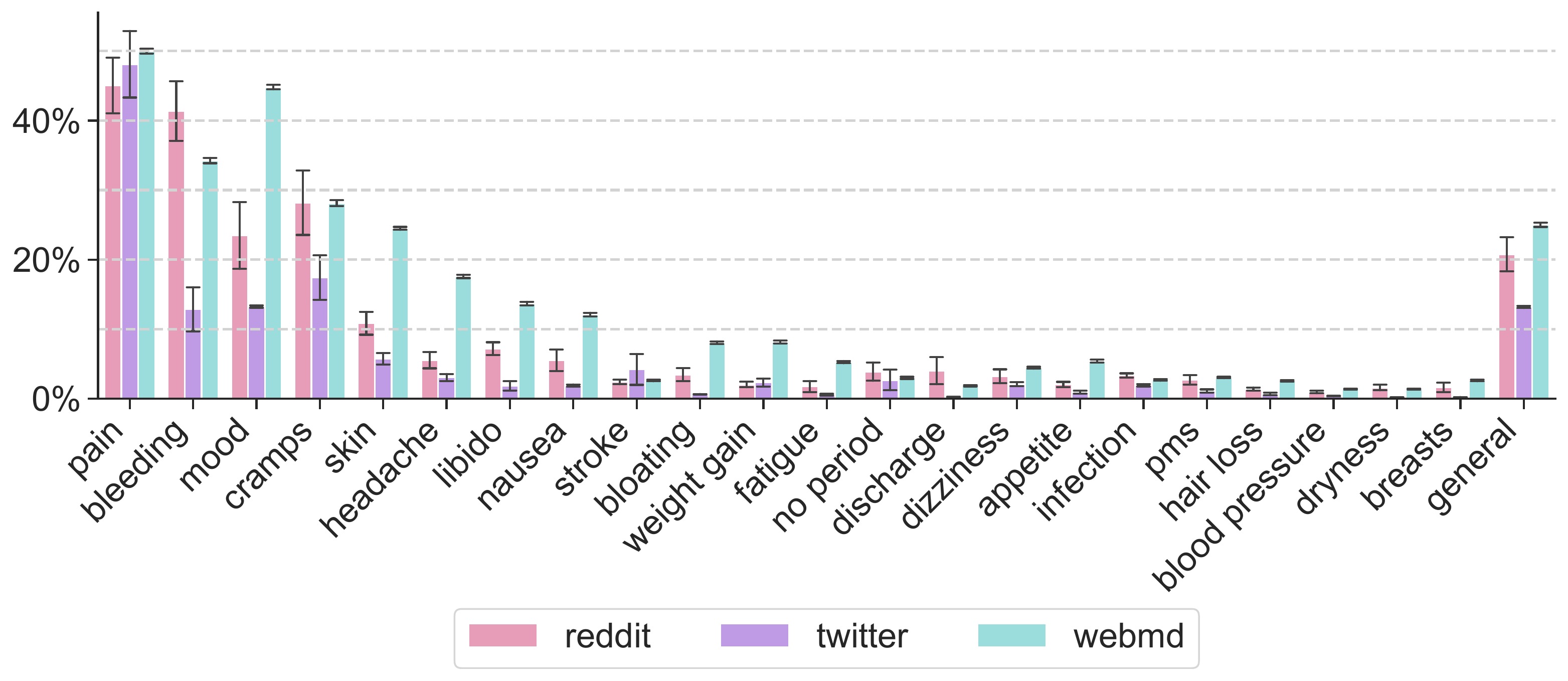}
    \includegraphics[width=0.49\textwidth]{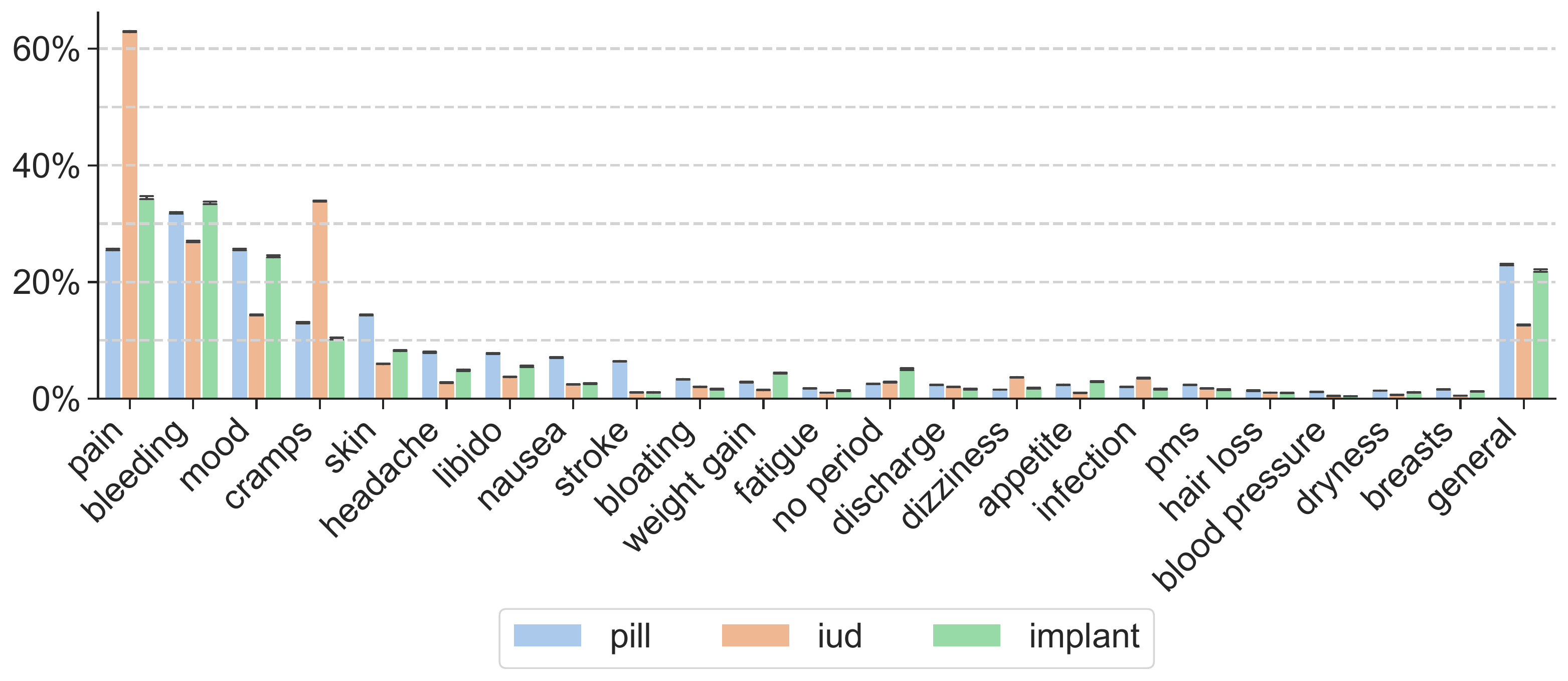}
    \caption{Distribution of the side effect mentions across the platforms and methods. Bars represent the percents of documents for the specified platform or method mentioning \textit{any} side effect that also mention the \textit{specified} side effect. 
    See Table \ref{table:lexicon-coverage} for the denominator sizes. 
    Error bars indicate the standard deviation over 20 bootstrapped samples of the datasets.
    } 
    \label{figure:barplot-side-effects-distribution}
\end{figure*}

\begin{figure}[t]
    \centering

        \centering
        \includegraphics[width=0.49\linewidth]{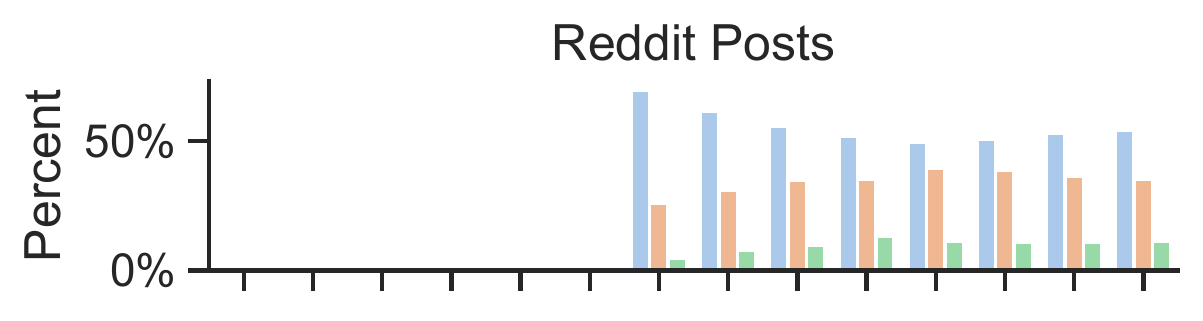}
        \includegraphics[width=0.49\linewidth]{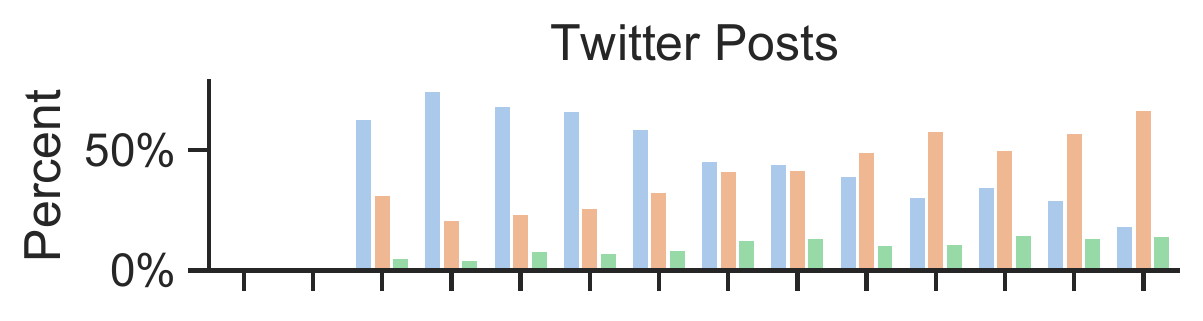}
        \includegraphics[width=0.49\linewidth]{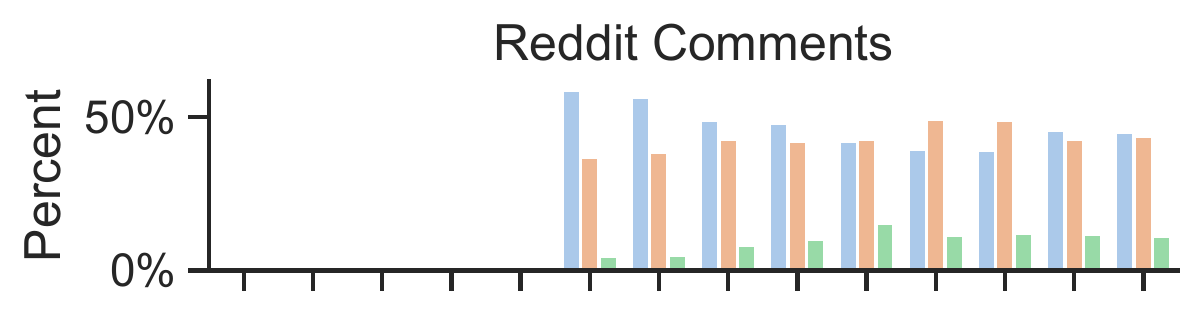}
        \includegraphics[width=0.49\linewidth]{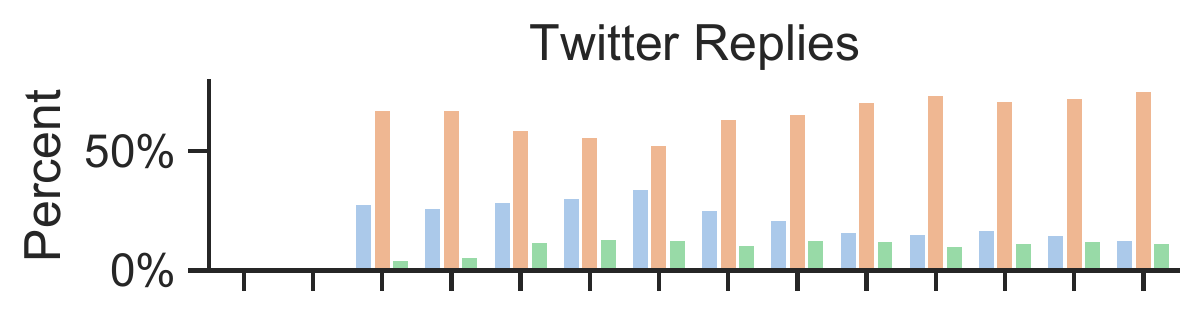}
        \includegraphics[width=0.49\linewidth]{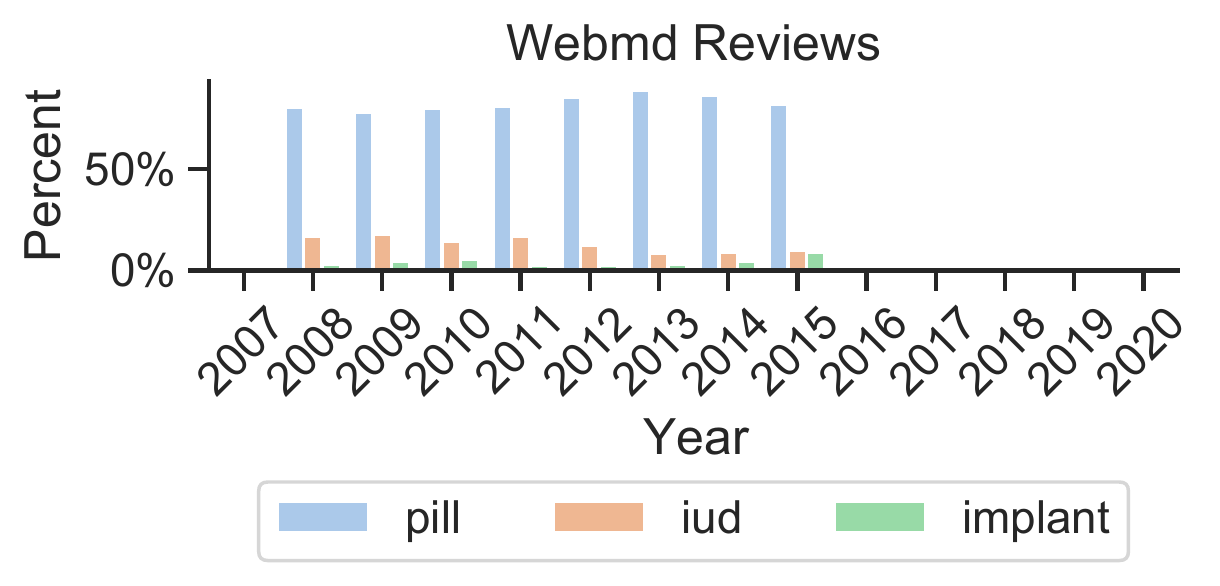}
    
    \caption{Document distributions of methods over time.}
    
    \label{figure:bc-type-distributions}
\end{figure}

\begin{figure}[!ht]

    \centering
    
    \includegraphics[width=0.1805\textwidth]{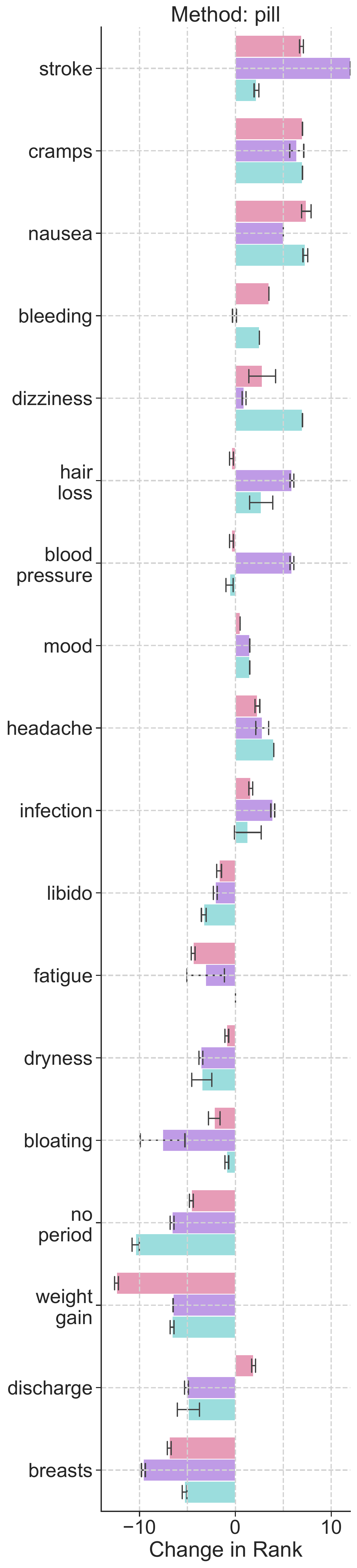}
    \includegraphics[width=0.135\textwidth]{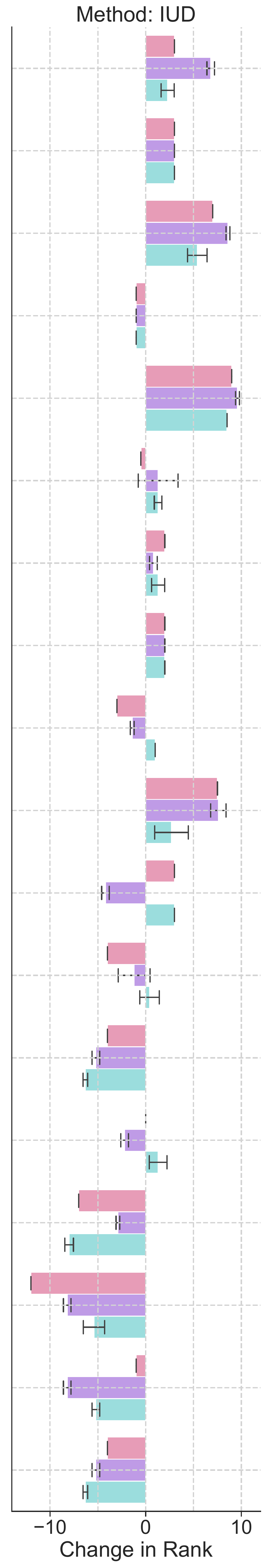}
    \includegraphics[width=0.135\textwidth]{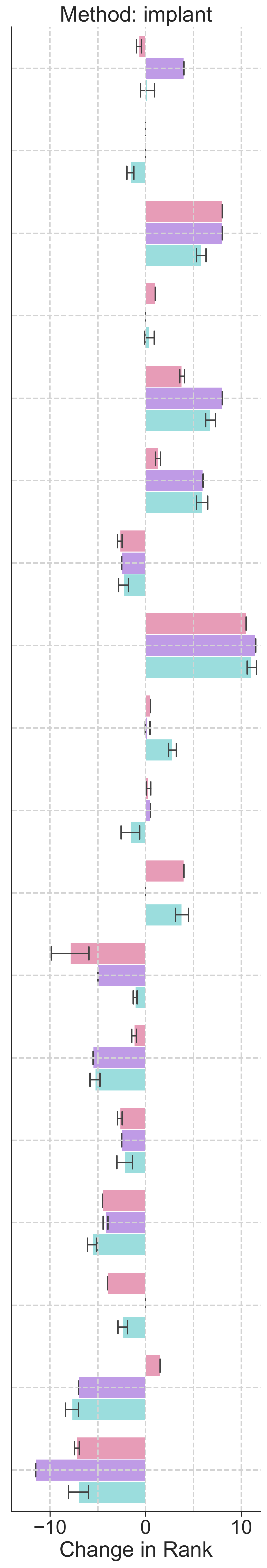}
    \\
    \includegraphics[width=0.25\textwidth]{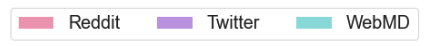}
    
    \caption{Distribution of the side effect mentions across the platforms. Bars represent the \textbf{differences} between the rank reported in the survey results from \citet{nelson2018women} and the rank on the specified platform. Platform ranks are determined by first finding the frequency of side effect mentions; these are the percent of documents mentioning \textit{any} side effect that also mention the \textit{specified} side effect. See Table \ref{table:lexicon-coverage} for the denominator sizes. Bars to the \textbf{right} of the x-axis represent side effects that are mentioned \textbf{more frequently} on the platforms than are reported in the survey.}
    
    \label{figure:barplot-side-effects-distribution-comparison}
    
\end{figure}

\begin{figure*}[!ht]

    \centering
    \begin{subfigure}[b]{0.24\textwidth}
        \includegraphics[width=\textwidth]{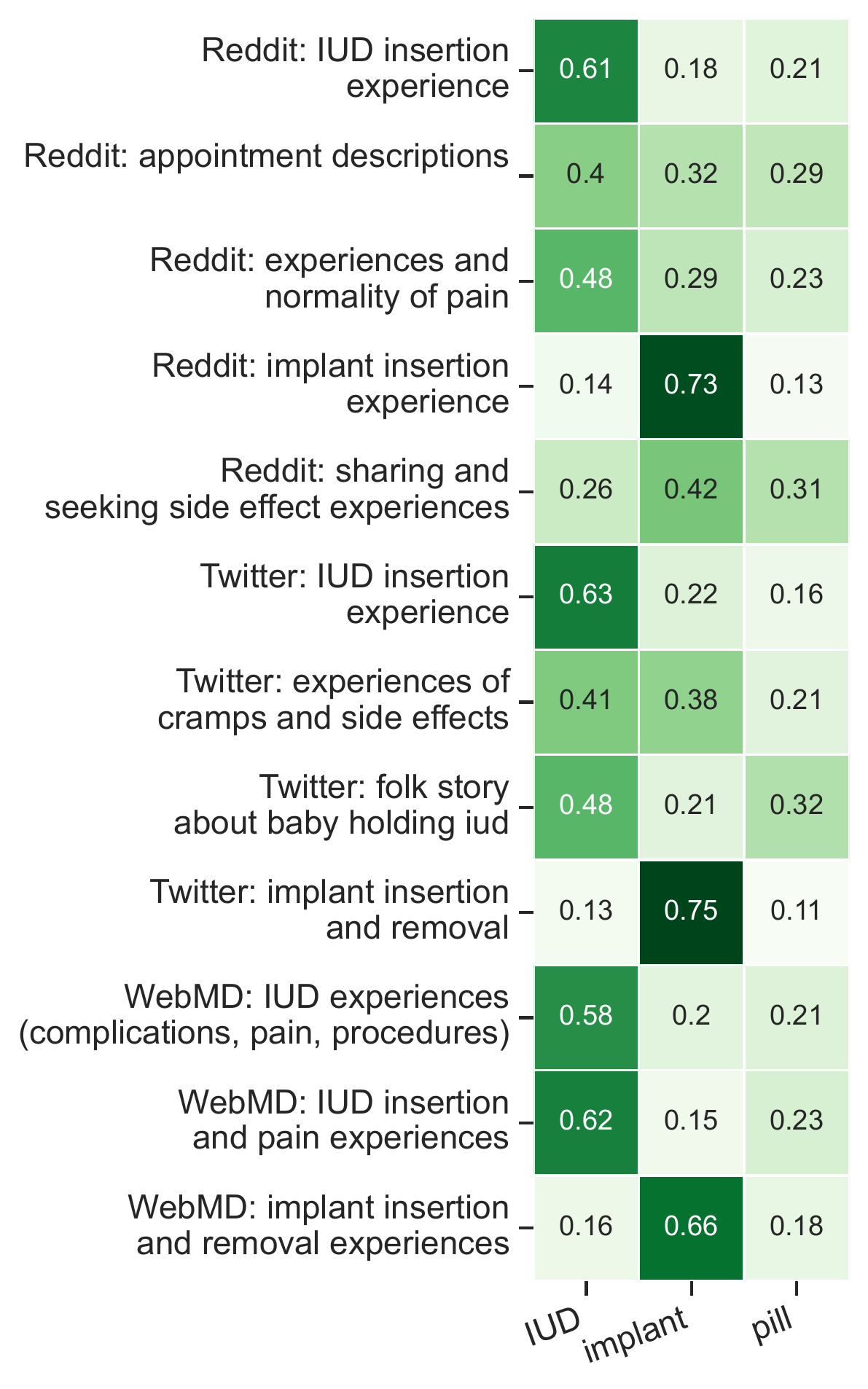}
        \caption{Storytelling}
    \end{subfigure}
    \begin{subfigure}[b]{0.24\textwidth}
        \includegraphics[width=\textwidth]{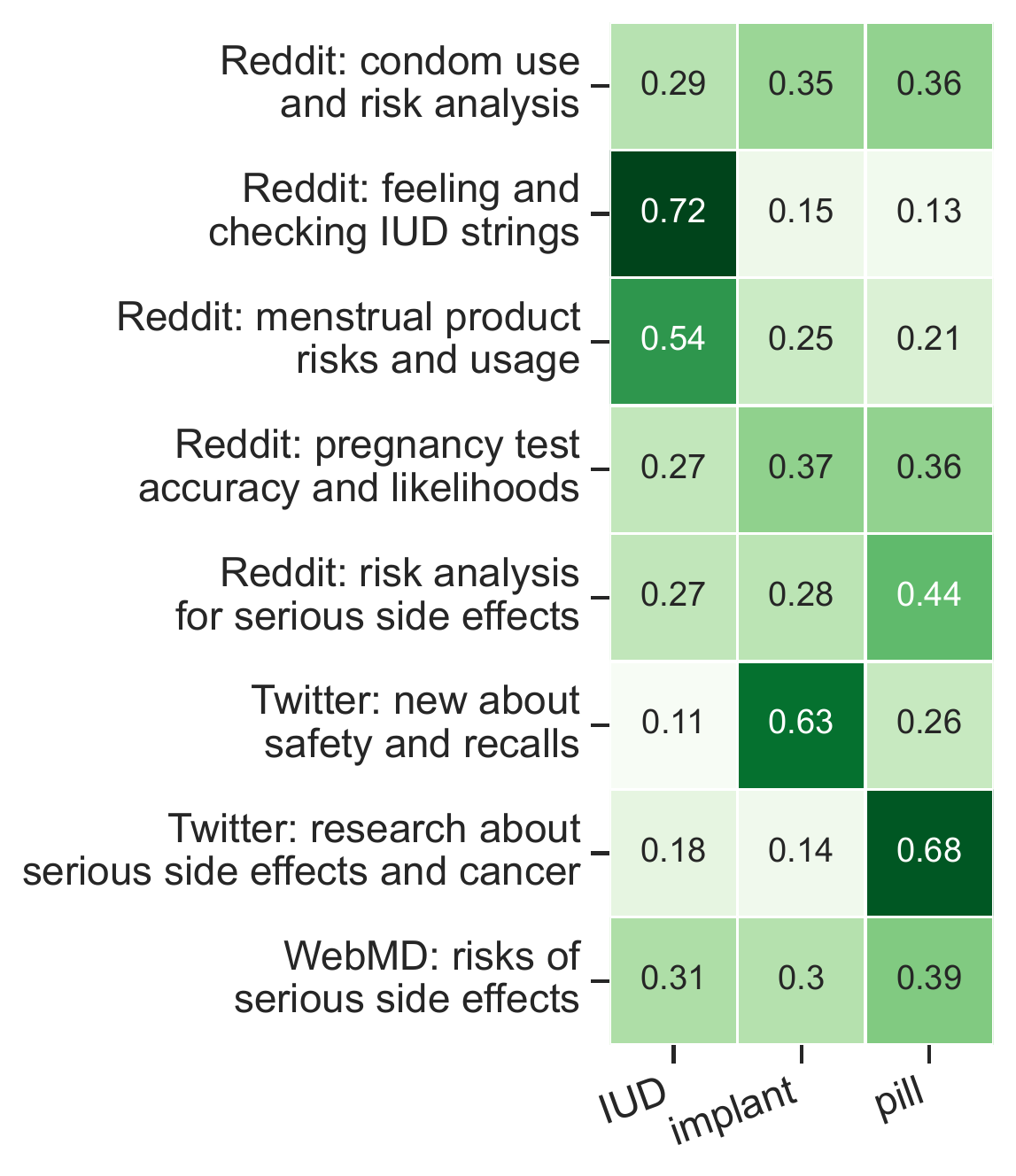}
        \caption{Risk Analysis}
    \end{subfigure}
    \begin{subfigure}[b]{0.24\textwidth}
        \includegraphics[width=\textwidth]{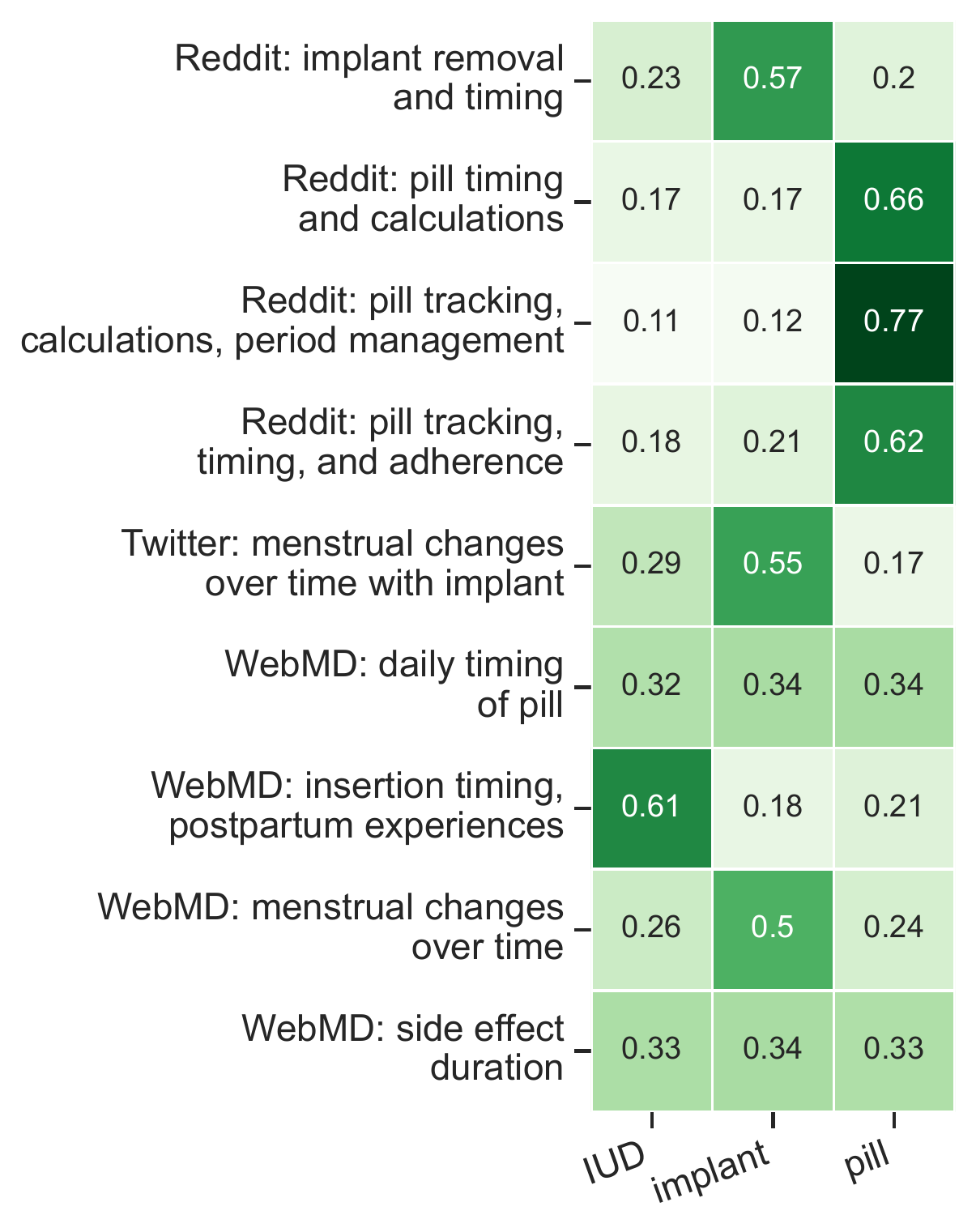}
        \caption{Timing \& Calculations}
    \end{subfigure}
    \begin{subfigure}[b]{0.24\textwidth}
        \includegraphics[width=\textwidth]{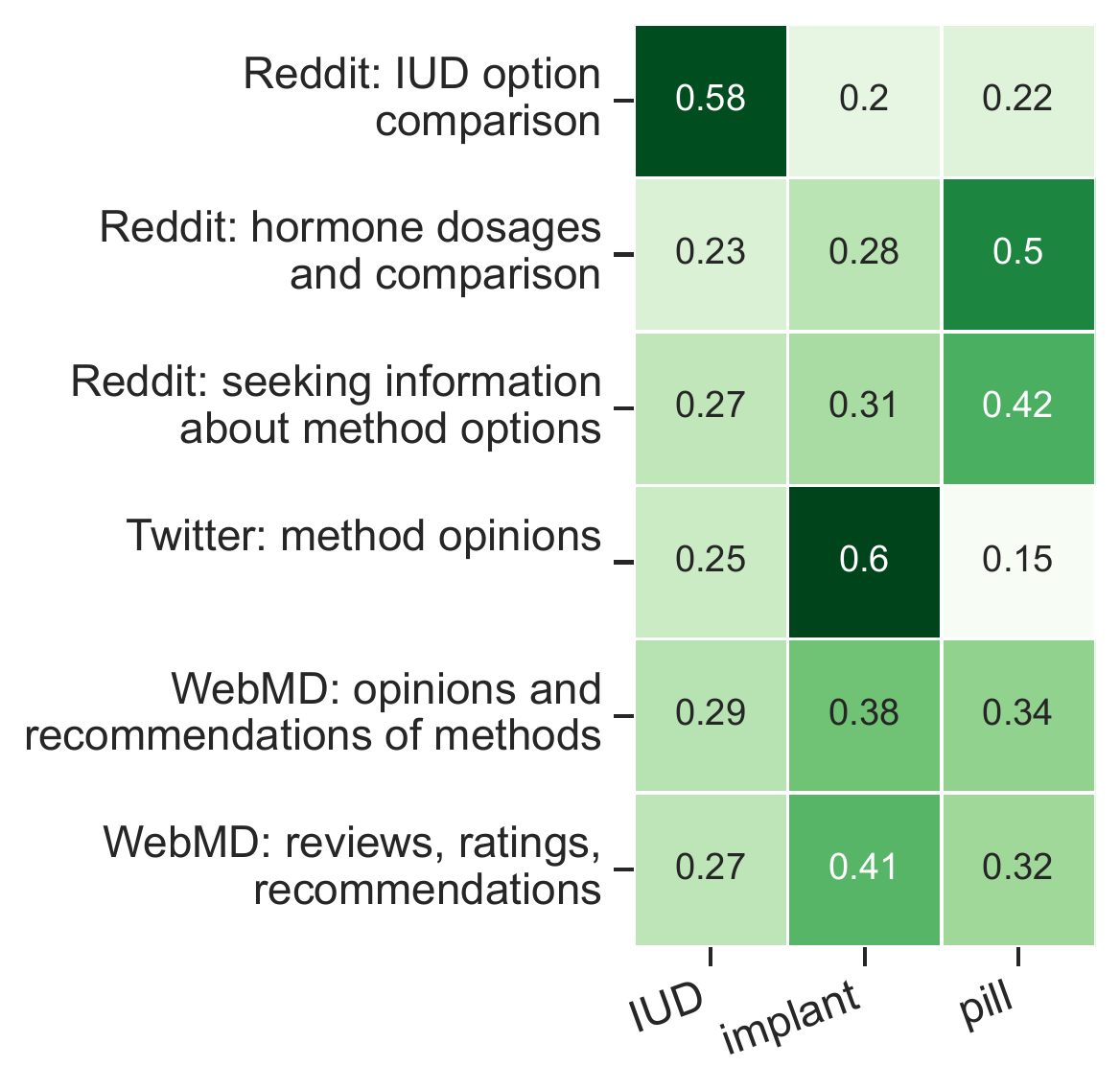} 
        \caption{Method \& Dose Comparison}
    \end{subfigure}

    \caption{Mean topic probabilities across methods for four of our sensemaking clusters. Rows are normalized to sum to one.}
    
    \label{figure:heatmaps-topics-methods}
    
\end{figure*}

\subsection{Sensemaking Topic Model}

We compare sensemaking activity distributions through an unsupervised, bottom-up analysis using topic modeling.
Topic modeling is an automatic method to identify prominent themes and discourses in a dataset \citep{blei2003latent} and is a popular unsupervised technique for the analysis of online health communities \citep{nobles2018std,abebe2020quantifying}.
While some sensemaking themes might be shared across settings, other themes might be more prevalent in certain platforms and side effects.
This process helps us to avoid biases in our data coding (which might overlook certain themes while over-representing others) by quickly revealing frequent themes in each dataset \citep{nelson2020computational}.

\subsubsection{Model training.}

We trained a latent Dirichlet allocation (LDA) topic model \citep{blei2003latent} on each of our datasets, by combining the Reddit posts and comments and the Twitter posts and replies for training, resulting in three models, using MALLET\footnote{\url{http://mallet.cs.umass.edu/}} for training.
LDA remains a highly consistent and reliable model \citep{harrando-etal-2021-apples,hoyle2022AreNT}, especially when trained via Gibbs sampling for smaller datasets.
We balanced each training set, sampling 8,000 documents for each method for each of the Reddit and Twitter datasets and 800 documents for each method for the WebMD dataset (see Figure \ref{figure:bc-type-distributions} for method distributions).
By balancing the training set, we avoid weighting the topics toward a certain method.

We removed a set of frequent stopwords,
numbers, punctuation, and duplicate documents from the training sets.
Removing stopwords and duplicate documents has been shown to improve the legibility of the final topics, whereas stemming can be harmful \citep{schofield-mimno-2016-comparing,schofield-etal-2017-pulling,schofield-etal-2017-quantifying}.
To avoid capitalization discrepancies,
we lowercase all text.
We experimented with different numbers of topics and found 35 to be interpretable across the datasets; at this number of topics, we observed topics that were neither too fine-grained (e.g., splitting a single theme across multiple topics) nor too high level (e.g., combining themes that should be separated), and that produced reasonable evaluation scores (see below).
However, we emphasize that there is no ``correct'' number of topics and that this method is used for exploration and interpretation.

\subsubsection{Model evaluation.}
\label{topic-model-evaluation}

While we do not have space here to list the full sets of topics for each dataset, we provide the topics, most probable words, and example paraphrased documents in our code repository for manual examination.
We report human evaluation of our topics following the recommendations in \citet{hoyle2021automated}.
Using the \textit{word intrusion task} \citep{chang2009reading}, we show two expert annotators (the first two authors) and one non-expert annotator a set of the four most probable words plus an ``intruder'' word that has low probability for the current topic but high probability for another topic.
We report the proportion of topics for which the annotators identified the intruder.

We found that our Reddit and WebMD topics have performance much higher than a random baseline of 0.2 (annotator accuracies for Reddit: 0.71, 0.74, 0.77, WebMD: 0.54, 0.66, 0.77) while the Twitter topics have lower performance but are still substantially above the random baseline (Twitter: 0.46, 0.46, 0.51).
The first (non-expert) annotator consistently had lower scores.
The lower performance on Twitter is expected, as text processing methods are notoriously challenged by the short text lengths and non-standard language \citep{gimpel-etal-2011-part}, and the short tweets require contextual knowledge to interpret.
This vulnerability of the word intrusion task to esoteric topics is noted by \citet{hoyle2021automated}.

We also calculate the ``UMass'' Coherence: the log probability that a document containing at least one instance of a higher-ranked word also contains at least one instance of a lower-ranked word \citep{roder2015exploring}.
We find the highest mean coherence scores across topics for Reddit ($-416$), lower scores for WebMD ($-577$), and lowest scores for Twitter ($-712$).
We note the criticisms of this and other automatic metrics in \citet{hoyle2021automated}; in comparison to human evaluation, automatic metrics can exaggerate differences between models.

\subsubsection{Identifying sensemaking themes.}

Across the datasets, we find that many of the topics discuss birth control methods (including one method in their 10 most probable words), side effects, pregnancy, and access (costs, appointments).
We then identify a series of cross-cutting sensemaking topics.
These topics include discussions of information seeking/sharing, educational links and resources, how-to explanations, experience seeking/sharing, emotional support, and other sensemaking-related discussions.
Two researchers independently coded the topics as more or less related to sensemaking.
The researchers then conferred and agreed upon a final set of topics and assigned them to thematic clusters.
During this coding, we relied on the sensemaking definition from \citet{andalibi2021sensemaking}: ``\textit{sensemaking describes how individuals make sense of complex phenomena by constructing mental models that draw on new or existing experiences, information, emotions, ideas, and memories}.''
We also drew inspiration from prior work studying sensemaking in online healthcare communities, particularly those works also focused on reproductive healthcare (see \S Related Work).
We further explore and validate these sensemaking topics by hand-labeling a small subsection of the data with social support goals \citep{yang2019seekers}.
After coding 150 documents for each dataset, we measured the agreement between the annotators using Krippendorff $\alpha$, as each document could receive zero, one, or more labels.
Our agreement was acceptable, with a score across the labels of 0.74.\footnote{Lower scores are not surprising for subjective language labeling tasks \citep{artstein2008inter,godwin2016collecting}, and our scores are substantially higher than the agreement scores for a very similar classification task in \citet{rivas2020classification}.}

To compare the topics across the platforms, we aligned the topics from the different models using Jensen-Shannon divergence (JSD) for the word distributions associated with each pair of topics. 
JSD is frequently used in prior work to compare topic distributions \citep{hall-etal-2008-studying,fang2012mining}.
After manual examination of the ranked topic pairs, we categorized topics with JSD scores below 0.6 as aligned across the datasets, and those with scores above 0.8 we considered diverging.

%% file: tables/lexicon_coverage.tex
\begin{table}[t]
    \scriptsize
    \begin{center}
    \begin{tabular}{p{1.9cm}p{2.5cm}p{2.7cm}}
    \hline
    \textbf{Source} & \textbf{\# Texts with Side Effect} & \textbf{\% Texts with Side Effect} \\
    \hline
    \textbf{Reddit Posts}
    & 53,027 / 72,731
    & 73\%
    \\
    \textbf{Reddit Comments}
    & 119,780 / 238,568
    & 50\%
    \\
    \textbf{Twitter Posts}
    & 61,698 / 513,017
    & 12\%
    \\
    \textbf{Twitter Replies}
    & 47,049 / 244,140
    & 19\%
    \\
    \textbf{WebMD Reviews}
    & 16,176 / 18,110
    & 89\%
    \\
    \hline
    \end{tabular}
    \end{center}
    \caption{The coverage of the side effects lexicon.}
    \label{table:lexicon-coverage}
\end{table}

%% file: sections/5-results.tex
\section{Results}

\subsection{Results: Birth Control Methods Across Platforms}

We compare the rates of discussion of the different birth control methods across platforms and over time.
We find that the types of questions that users ask and the type of information they share differs based on birth control method.
We measure the rise and fall of discussions of different birth control methods over time, and we compare these distributions across platforms.
Importantly, these patterns do not necessarily indicate real-world increases or decreases in use of different methods, but they do show the concerns and interests of users on these platforms.

Figure \ref{figure:bc-type-distributions} shows the frequency of documents for each medication method by platform over time. 
The pill is the most popular method in both \textit{Reddit Posts} and \textit{WebMD Reviews}, while the pill and IUD are tied in \textit{Reddit Comments}.
Discussions of the pill on WebMD are much more frequent that discussions of other birth control methods and do not show a rise in interest for the IUD, counter to both the other datasets and our original hypothesis.
On Twitter, the pill begins as the most popular but is replaced by the IUD in posts, while replies always center on the IUD.
Twitter replies also increase sharply after 2016, differing from the Twitter posts (so is unlikely to be a symptom of our keywords).
This suggests that the IUD generates more discussion on Twitter, especially post-2016, compared to the other birth control methods.
On Reddit, the number of posts discussing the pill and IUD are similar, with slightly more posts discussing the pill, though this difference is erased in the comments.
The implant is always the least discussed of the three methods.

\subsection{Results: Side Effects Across Platforms}

Figure \ref{figure:bc-type-distributions} shows the distributions of the side effects across the different platforms, where frequency proportions are calculated by dividing the number of texts mentioning the specified side effect by the total number of texts mentioning any side effect (i.e., $p(s_i|a)$ where $s_i$ is the specific side effect and $a$ is any side effect).
Discussions of different methods have changed dramatically over time on Twitter, with discussions of the IUD rising and the pill falling.
The pill remains most most mentioned in Reddit posts.

Figure \ref{figure:barplot-side-effects-distribution} shows the distribution of side effects across platforms.
We find that pain, cramps, menstrual bleeding irregularities, mood changes, and skin conditions are the most commonly discussed side effects.
In particular, \textit{pain is consistently and frequently discussed across the platforms} and is an outlier among the side effects.
Discussions of stroke are more frequent in Twitter posts than in the other datasets, but for all other side effects, Twitter has the lowest discussion frequencies (perhaps because of stigma around publicly sharing such information compared to the more frequently pseudonymous settings on Reddit and WebMD).
Except for pain, menstrual bleeding, and severe effects, WebMD has the highest frequency of side effects discussions compared to the other platforms.
In particular, WebMD has higher frequencies for discussions of mood changes, skin conditions, and headache discussions.
Reddit has has the highest frequency of discussions for menstrual bleeding irregularities.

Figure \ref{figure:barplot-side-effects-distribution} also compares the side effects by birth control method.
Across the birth control methods, we again find that pain is a frequently mentioned side effect, but it is most frequently mentioned in discussions that also mention the IUD rather than the pill or implant.
The implant is the only method shown to cause weight gain \citep{barr2010managing}, so this association is expected---but the discussions of weight gain for the other methods are less expected and could indicate that this potential side effect is a concern across methods.
Menstrual bleeding, mood changes, headache, and libido are least often discussed in with the IUD, as are general mentions of side effects,  while nausea, stroke, and skin conditions are most often discussed in discussions of the pill.

\subsection{Results: Comparison to Observed Distributions}

Figure \ref{figure:barplot-side-effects-distribution-comparison} shows the differences between our observed distributions of side effect mentions and the reported distributions of side effect experiences in \citet{nelson2018women} (described above).
We find large differences between how frequently side effects are discussed online compared with how frequently they are reported in the survey from \citet{nelson2018women}.
For example, strokes are rarely experienced according to \citet{nelson2018women}, but when mentioned with the pill, they are discussed more frequently on Twitter in comparison to other side effects; this is unsurprising, since sensational topics are frequently discussed on Twitter.
Mood changes are more likely to be discussed across the platforms for the implant in comparison to the survey data.
Dizziness is more likely to be discussed for the IUD than expected, while weight gain is universally discussed less frequently than expected.
These patterns indicate cases in which users turn to the internet more or less frequently than expected based on their self-reported experiences in the survey.

\subsection{Results: Sensemaking Across Platforms}

Our final set of sensemaking themes included: \textbf{storytelling} (e.g., \textit{implant insertion and removal} on Twitter), \textbf{risk analysis} (e.g., \textit{risks of serious side effects} on WebMD), \textbf{timing and calculations} (e.g., \textit{daily timing of pill} on WebMD), \textbf{method and dose comparison} (e.g., \textit{hormone dosages and comparison} on Reddit), \textbf{causal reasoning} (e.g., \textit{weight changes and causes} on Reddit), and \textbf{information and explanations} (e.g., \textit{research on the male pill} on Twitter).
We show four of these themes (others omitted for space) in Figure \ref{figure:heatmaps-topics-methods}, with the topics from each platform included in that theme and their probabilities for each birth control method.
We observe a greater number of storytelling topics for the IUD and implant than for the pill, and examination of these topics shows that insertion and removal experiences fuel this pattern.
Each method is associated with risks and is included in method comparisons, but the pill in particular is included in discussions of timing and calculations.

Using Jensen-Shannon divergence to compare topics across the platforms, we find that the most \textbf{aligned} ($JSD < 0.6$) topic categories were: \textit{weight changes}, \textit{general side effects}, \textit{IUD insertion experiences}, \textit{implant insertion experiences}, \textit{menstrual timing and cycles}, and \textit{bleeding changes}.
Each of our three datasets included at least one representative topic from these categories.

The most \textbf{diverging} ($JSD > 0.8$) topics were: \textit{causes and side effects of vaginal infections} and \textit{explanations of how birth control works} on Reddit; \textit{IUD jokes and random}, \textit{viral folk stories}, and \textit{implant news about unexpected experiences} on Twitter; and \textit{treating or causing other medical issues} and \textit{secondhand advice and experiences} on WebMD.
These topics were the most unique to their training dataset, without directly comparable topics in the other datasets.

%% file: sections/8-discussion.tex
\section{Discussion}

 \paragraph{Methods and side effects across platforms.}
In response to our first research question, we find that birth control discussions on Twitter, WebMD, and Reddit substantially differ in their distributions of methods and side effect mentions.
We cannot claim that using a method or experiencing a side effect \textit{cause} people to choose a specific platform, but our observations add detail and sometimes contradict prior findings.
For example, in a study of general healthcare information seeking, \citet{choudhury2014seeking} found that people more often use search engines for serious and stigmatized conditions and more often use Twitter to discuss symptoms.
We do not include search engines in our study, but in our comparison across platforms, we find that Twitter has a higher frequency of severe side effect discussions for birth control, while WebMD and Reddit have higher frequencies of general side effect discussions.
Our focus on birth control might explain these variations, as many of the birth control side effects are themselves highly stigmatized (e.g., menstrual bleeding, vaginal discharge). 
The discussion of severe side effects on Twitter is likely related to Twitter's tendency toward sensational content, and could also explain the relative frequency of IUD discussion on Twitter; the IUD has been associated with both severe side effects and potential legal bans
\citep{nobles2019repeal}.
These findings emphasize the importance of considering platform setting when studying online patterns related to specific health conditions, medications, and side effects.

\paragraph{Online sensemaking practices of birth control users.}
In response to our second research question, our identification and comparison of topics that align with sensemaking themes reveals important activities specific to birth control and to the different platforms and methods.
Prior work has found that online communities focused on reproductive healthcare (e.g., pregnancy, vulvodynia) employ strategies related to validation \citep{andalibi2021sensemaking}, information management \citep{patel2019feel,young2019girl,andalibi2021sensemaking}, personal tracking \citep{chopra2021living}, and identifying causation \citep{patel2019feel}.
We found these themes again in our birth control communities, with variations; for example, personal tracking is also a prominent theme in birth control discussions, but it is focused on pill timings and calculations and on self-observations of side effects. 
Identifying causation is a common concern for those experiencing infertility \citep{patel2019feel}, and we find this theme again in our datasets but focused on side effects like weight gain.
This combination of themes reflects the unique problems facing birth control users, as they choose between ``least bad'' options, struggle to identify and treat side effects, determine risk of pregnancy given their circumstances, and avoid rare but alarming outcomes like stroke and heart attack.

\paragraph{Storytelling, pain, and the IUD.}
We would like to particularly highlight storytelling as an important sensemaking strategy for birth control users.
Storytelling is known to help communities work through trauma \citep{tangherlini2000heroes}, but birth control users employ storytelling specifically to address physical pain.
Pain is a major cross-cutting theme across the platforms; it is the most frequently and consistently discussed side effect, and it is most often mentioned alongside the IUD.
While \citet{barr2010managing} identifies pain as the second most common side effect (after bleeding changes) and \citet{dickerson2013satisfaction} identifies pain as the most common side effect for the IUD and second most common side effect for the implant, we find a much larger gap between pain and the next most frequent side effects in our analysis.
Pain is mentioned frequently in posts about the IUD and in posts whose probable topics are about seeking and sharing IUD insertion stories.
Pain is inherently a subjective experience that cannot be precisely communicated \citep{scarry1987body}, but sharing of personal experiences provides one way for a community to build a sense of what is \textit{normal} or \textit{to be expected}  \citep{patel2019feel,andalibi2021sensemaking}. 
We suggest that there is an urgent unmet need for (a) honest education and preparation before IUD insertions and (b) pain treatment options during this procedure.

\paragraph{Online discussions differ from survey reports.}
In comparison to the survey results in \citet{nelson2018women}, we find not only large differences in the frequency at which different side effects are discussed but also differences across the platforms.
It could be that the differences from the survey are due to the demographic distribution opting into the survey versus those opting to post online; future work surveying the demographic distribution of these users would address this question, but our results take a first step at measuring differences between these settings. 
We find, e,g., while mood changes are discussed at similar frequencies in comparison to the survey data across the platforms, other side effects like strokes, bloating, fatigue, bleeding, and dizziness might be discussed more, less, or at equal frequency ranks to the reports in \citet{nelson2018women}, depending on the platform.
Bloating is less frequently mentioned on all the platforms and for all the methods in comparison to the survey data, perhaps indicating a general lack of concern about this side effect in contrast to its reported frequency.
But bloating is much less frequent on Twitter for the pill and implant, which could also indicate that this particular setting is less suited to discussing this side effect, perhaps because of the embarrassing or intimate nature of this side effect.

\paragraph{Stigma, privacy, and contextual disclosures.}
Birth control can be a controversial, stigmatized, and intimate topic.
This can lead birth control users to seek out additional information privately.
For example, in a set of interviews of young Black and Hispanic women, \citet{yee2010role} found that a greater number reported seeking decision-marking support on the internet, citing its privacy, in comparison to other sources of information (e.g., talking to physicians, reading provided information).
Interpreting side effects, analyzing the risk of pregnancy, or normalizing a painful experience require disclosing personal details and stories.
Birth control users might analyze the risk and benefit of making these disclosures in certain settings and to certain audiences.
While social penetration theory \citep{altman1973social} posits that more disclosures are possible as social bonds deepen, prior work has also found that intimate language can be frequent among both close connections and strangers (but not in between) \citep{pei-jurgens-2020-quantifying}.
The variations we observe across methods, side effects, and sensemaking practices could be indicative of platform affordances for privacy, audience size, and anonymity, each of which can affect decisions to self-disclose.
For example, we found that side effect discussions are less frequent on Twitter, where users are facing a much larger and non-specialized audience, unlike the other platforms.
Giving users more platform-specific tools to control their audience size and membership \citep{,mondal2014understanding} could allow for more productive discussions of this sensitive topic.

\paragraph{Other factors influencing platform decisions.}
It is important to note that decisions to seek information online can correlate with demographic characteristics.
For example, in a survey of U.S. young adults, those with a sexual risk history (early sexual activity, involvement in an unintended pregnancy) less frequently reported using the internet as a source and more frequently reported seeking information from a doctor/nurse, and men more frequently reported using the internet than women \cite{khurana2015young}.
It is also possible that users follow a \textit{birth control journey}, where different needs at different points in one's journey can lead one to different platforms, as has been reported for other healthcare topics \citep{sannon2019really,andalibi2018announcing}.
These journeys can intersect with methods; for example, while the pill is a popular first method, many people report switching to the IUD as they gain more experience with birth control \citep{nelson2018women}.
This would mirror the journeys of those with invisible chronic illnesses who move from one platform to another as their needs evolve and as they grow more comfortable with self-disclosure \citep{sannon2019really}.
This is mirrored in intra-community research that models user trajectories in online cancer support groups, finding that users often transition from information-seeking to information-sharing roles over time \citep{yang2019seekers}.

\paragraph{Recommendations.}
Our results show that patterns found in one platform are not necessarily replicable on other online platforms, even when grounded in the same health topic.
We recommend that social media researchers compare results across multiple platforms to increase confidence in shared patterns.
If relying on a single platform, its choice should be guided by an understanding of which platforms might be more less or conducive to discussions of the desired methods, side effects, or sensemaking activities.
Our work also highlights benefits of mixing computational tools like topic modeling with lexicon-based methods and hand-annotation, and how these methods can be used to characterize difficult-to-identify themes like sensemaking practices; we recommend taking this careful approach, putting unsupervised results in context with fine-grained measurements.

%% file: sections/9-ethics.tex
\section{Broader Impacts, Ethics, and Limitations}
\label{section:ethics}

Our study was considered exempt under our institution's IRB.
However, while Reddit, Twitter, and WebMD posts and replies are ``public,'' they can contain highly personal information,
requiring a balance between potential harms and potential benefits to the community, as described in the guiding principles of the Belmont Report: \textit{respect for persons}, \textit{beneficence}, and \textit{justice}.\footnote{\url{https://www.hhs.gov/ohrp/regulations-and-policy}}
Considering possible harms, e.g., re-identification of those using stigmatized medications, we do not collect any user-specific information, and we do not infer medical conditions for individual users; instead, we rely on patterns averaged across many users.
We also anonymize and paraphrase any direct quotations.
To protect users' agency to edit or delete their data at its original source, we release our data collection lexicons but not copies of the collected data.
We balance these concerns and protective actions against the benefits of this research; among other benefits, our work highlights the unmet pain treatment needs of a population known to be discriminated against by physicians \citep{Samulowitz2018BraveMA} and examines the kinds of support needed by those facing difficult healthcare choices.

%% file: sections/10-limitations.tex
\section{Limitations}

Social media does not necessarily represent offline events, 
and unlike the surveys we use for comparison, we cannot control for demographics.
We study texts written in English and platforms that attract a U.S. audience, as the authors are all most familiar with this setting.
We focus on one specific community out of many, and we do not expect that the findings in this paper will necessarily generalize outside of the U.S. or to other online spaces; 
indeed, one of our results indicates that different platforms display different patterns.
Finally, averaging over posts allows us to track patterns and make comparisons but can also reduce nuance.
Our work is best read in conjunction with ethnographic studies like \citet{homewood2017turned} and \citet{daley2014influences}, which highlight individual voices of birth control users.

%% file: main.bbl
\begin{thebibliography}{59}
\providecommand{\natexlab}[1]{#1}

\bibitem[{Abebe et~al.(2020)Abebe, Giorgi, Tedijanto, Buffone, and
  Schwartz}]{abebe2020quantifying}
Abebe, R.; Giorgi, S.; Tedijanto, A.; Buffone, A.; and Schwartz, H. A.~A. 2020.
\newblock Quantifying Community Characteristics of Maternal Mortality Using
  Social Media.
\newblock In \emph{Proceedings of WWW}.

\bibitem[{Altman and Taylor(1973)}]{altman1973social}
Altman, I.; and Taylor, D.~A. 1973.
\newblock \emph{Social penetration: The development of interpersonal
  relationships.}

\bibitem[{Andalibi and Forte(2018)}]{andalibi2018announcing}
Andalibi, N.; and Forte, A. 2018.
\newblock Announcing Pregnancy Loss on Facebook: A Decision-Making Framework
  for Stigmatized Disclosures on Identified Social Network Sites.
\newblock In \emph{Proceedings of CHI}.

\bibitem[{Andalibi and Garcia(2021)}]{andalibi2021sensemaking}
Andalibi, N.; and Garcia, P. 2021.
\newblock Sensemaking and Coping After Pregnancy Loss: The Seeking and
  Disruption of Emotional Validation Online.
\newblock In \emph{Proceedings of CSCW}.

\bibitem[{Artstein and Poesio(2008)}]{artstein2008inter}
Artstein, R.; and Poesio, M. 2008.
\newblock Inter-coder agreement for computational linguistics.
\newblock \emph{Computational Linguistics}.

\bibitem[{Barr(2010)}]{barr2010managing}
Barr, N.~G. 2010.
\newblock Managing adverse effects of hormonal contraceptives.
\newblock \emph{American Family Physician}.

\bibitem[{Bietti, Tilston, and Bangerter(2019)}]{Bietti2019StorytellingAA}
Bietti, L.~M.; Tilston, O.; and Bangerter, A. 2019.
\newblock Storytelling as Adaptive Collective Sensemaking.
\newblock \emph{Topics in Cognitive Science}.

\bibitem[{Blei, Ng, and Jordan(2003)}]{blei2003latent}
Blei, D.~M.; Ng, A.~Y.; and Jordan, M.~I. 2003.
\newblock Latent {D}irichlet allocation.
\newblock \emph{JMLR}.

\bibitem[{Carron-Arthur et~al.(2015)Carron-Arthur, Ali, Cunningham, and
  Griffiths}]{carron2015help}
Carron-Arthur, B.; Ali, K.; Cunningham, J.~A.; and Griffiths, K.~M. 2015.
\newblock From help-seekers to influential users: a systematic review of
  participation styles in online health communities.
\newblock \emph{JMIR}.

\bibitem[{Chang et~al.(2009)Chang, Gerrish, Wang, Boyd-Graber, and
  Blei}]{chang2009reading}
Chang, J.; Gerrish, S.; Wang, C.; Boyd-Graber, J.~L.; and Blei, D.~M. 2009.
\newblock Reading tea leaves: How humans interpret topic models.
\newblock In \emph{Advances in NeurIPS}.

\bibitem[{Chopra et~al.(2021)Chopra, Zehrung, Shanmugam, and
  Choe}]{chopra2021living}
Chopra, S.; Zehrung, R.; Shanmugam, T.~A.; and Choe, E.~K. 2021.
\newblock Living with Uncertainty and Stigma: Self-Experimentation and
  Support-Seeking around Polycystic Ovary Syndrome.
\newblock In \emph{Proceedings of CHI}.

\bibitem[{Daley(2014)}]{daley2014influences}
Daley, A.~M. 2014.
\newblock What influences adolescents' contraceptive decision-making? A
  meta-ethnography.
\newblock \emph{Journal of Pediatric Nursing}.

\bibitem[{Daniels and Abma(2020)}]{daniels2020current}
Daniels, K.; and Abma, J.~C. 2020.
\newblock Current Contraceptive Status Among Women Aged 15--49: United States,
  2017--2019.
\newblock \emph{NCHS Data Brief No. 388}.

\bibitem[{De~Choudhury, Morris, and White(2014)}]{choudhury2014seeking}
De~Choudhury, M.; Morris, M.~R.; and White, R.~W. 2014.
\newblock Seeking and Sharing Health Information Online: Comparing Search
  Engines and Social Media.
\newblock In \emph{Proceedings of CHI}.

\bibitem[{Dickerson et~al.(2013)Dickerson, Diaz, Jordan, Chirina, Goddard,
  Carr, and Carek}]{dickerson2013satisfaction}
Dickerson, L.~M.; Diaz, V.~A.; Jordan, J.; Chirina, S.; Goddard, J.~A.; Carr,
  K.~B.; and Carek, P.~J. 2013.
\newblock Satisfaction, early removal, and side effects associated with
  long-acting reversible contraception.
\newblock \emph{Family Medicine}.

\bibitem[{Fang et~al.(2012)Fang, Si, Somasundaram, and Yu}]{fang2012mining}
Fang, Y.; Si, L.; Somasundaram, N.; and Yu, Z. 2012.
\newblock Mining Contrastive Opinions on Political Texts Using
  Cross-Perspective Topic Model.
\newblock In \emph{Proceedings of ICWSM}.

\bibitem[{Genuis and Bronstein(2017)}]{genuis2017looking}
Genuis, S.~K.; and Bronstein, J. 2017.
\newblock Looking for “normal”: Sense making in the context of health
  disruption.
\newblock \emph{JASIST}.

\bibitem[{Gimpel et~al.(2011)Gimpel, Schneider, O{'}Connor, Das, Mills,
  Eisenstein, Heilman, Yogatama, Flanigan, and Smith}]{gimpel-etal-2011-part}
Gimpel, K.; Schneider, N.; O{'}Connor, B.; Das, D.; Mills, D.; Eisenstein, J.;
  Heilman, M.; Yogatama, D.; Flanigan, J.; and Smith, N.~A. 2011.
\newblock Part-of-Speech Tagging for {T}witter: Annotation, Features, and
  Experiments.
\newblock In \emph{Proceedings of ACL}.

\bibitem[{Godwin and Piwek(2016)}]{godwin2016collecting}
Godwin, K.; and Piwek, P. 2016.
\newblock Collecting reliable human judgements on machine-generated language:
  The case of the {QG-STEC} data.
\newblock In \emph{Proceedings of INGL}.

\bibitem[{Hall, Jurafsky, and Manning(2008)}]{hall-etal-2008-studying}
Hall, D.; Jurafsky, D.; and Manning, C.~D. 2008.
\newblock Studying the History of Ideas Using Topic Models.
\newblock In \emph{Proceedings of EMNLP}.

\bibitem[{Harrando, Lisena, and Troncy(2021)}]{harrando-etal-2021-apples}
Harrando, I.; Lisena, P.; and Troncy, R. 2021.
\newblock Apples to Apples: A Systematic Evaluation of Topic Models.
\newblock In \emph{Proceedings of RANLP}.

\bibitem[{Homewood and Heyer(2017)}]{homewood2017turned}
Homewood, S.; and Heyer, C. 2017.
\newblock Turned On / Turned Off: Speculating on the Microchip-Based
  Contraceptive Implant.
\newblock In \emph{Proceedings of DIS}.

\bibitem[{Hoyle et~al.(2021)Hoyle, Goel, Hian-Cheong, Peskov, Boyd-Graber, and
  Resnik}]{hoyle2021automated}
Hoyle, A.; Goel, P.; Hian-Cheong, A.; Peskov, D.; Boyd-Graber, J.; and Resnik,
  P. 2021.
\newblock Is Automated Topic Model Evaluation Broken? The Incoherence of
  Coherence.
\newblock \emph{Advances in NeurIPS}.

\bibitem[{Hoyle et~al.(2022)Hoyle, Goel, Sarkar, and Resnik}]{hoyle2022AreNT}
Hoyle, A.~M.; Goel, P.; Sarkar, R.; and Resnik, P. 2022.
\newblock Are Neural Topic Models Broken?
\newblock \emph{ArXiv}, abs/2210.16162.

\bibitem[{Khurana and Bleakley(2015)}]{khurana2015young}
Khurana, A.; and Bleakley, A. 2015.
\newblock Young adults’ sources of contraceptive information: variations
  based on demographic characteristics and sexual risk behaviors.
\newblock \emph{Contraception}.

\bibitem[{Mamykina, Nakikj, and Elhadad(2015)}]{mamykina2015collective}
Mamykina, L.; Nakikj, D.; and Elhadad, N. 2015.
\newblock Collective Sensemaking in Online Health Forums.
\newblock In \emph{Proceedings of CHI}.

\bibitem[{Manzer and Bell(2022)}]{manzer2022did}
Manzer, J.~L.; and Bell, A.~V. 2022.
\newblock “Did I Choose a Birth Control Method Yet?”: Health Care and
  Women’s Contraceptive Decision-Making.
\newblock \emph{Qualitative Health Research}.

\bibitem[{Merz et~al.(2020)Merz, Guti{\'e}rrez-Sacrist{\'a}n, Bartz, Williams,
  Ojo, Schaefer, Huang, Li, Sandoval, Ye et~al.}]{merz2020population}
Merz, A.~A.; Guti{\'e}rrez-Sacrist{\'a}n, A.; Bartz, D.; Williams, N.~E.; Ojo,
  A.; Schaefer, K.~M.; Huang, M.; Li, C.~Y.; Sandoval, R.~S.; Ye, S.; et~al.
  2020.
\newblock Population attitudes toward contraceptive methods over time on a
  social media platform.
\newblock \emph{American Journal of Obstetrics and Gynecology}.

\bibitem[{Mondal et~al.(2014)Mondal, Liu, Viswanath, Gummadi, and
  Mislove}]{mondal2014understanding}
Mondal, M.; Liu, Y.; Viswanath, B.; Gummadi, K.~P.; and Mislove, A. 2014.
\newblock Understanding and specifying social access control lists.
\newblock In \emph{Symposium On Usable Privacy and Security}.

\bibitem[{Nahata, Chelvakumar, and Leibowitz(2017)}]{nahata2017gender}
Nahata, L.; Chelvakumar, G.; and Leibowitz, S. 2017.
\newblock Gender-affirming pharmacological interventions for youth with gender
  dysphoria: when treatment guidelines are not enough.
\newblock \emph{Annals of Pharmacotherapy}.

\bibitem[{Nelson et~al.(2017)Nelson, Cohen, Galitsky, Hathaway, Kappus,
  Kerolous, Patel, and Dominguez}]{nelson2018women}
Nelson, A.~L.; Cohen, S.; Galitsky, A.; Hathaway, M.; Kappus, D.; Kerolous, M.;
  Patel, K.; and Dominguez, L. 2017.
\newblock Women's perceptions and treatment patterns related to contraception:
  results of a survey of US women.
\newblock \emph{Contraception}.

\bibitem[{Nelson(2020)}]{nelson2020computational}
Nelson, L.~K. 2020.
\newblock Computational grounded theory: A methodological framework.
\newblock \emph{Sociological Methods \& Research}.

\bibitem[{Nobles et~al.(2018)Nobles, Dreisbach, Keim-Malpass, and
  Barnes}]{nobles2018std}
Nobles, A.; Dreisbach, C.; Keim-Malpass, J.; and Barnes, L. 2018.
\newblock ``Is this an STD? Please help!'': Online information seeking for
  sexually transmitted diseases on {R}eddit.
\newblock In \emph{Proceedings of ICWSM}.

\bibitem[{Nobles, Dredze, and Ayers(2019)}]{nobles2019repeal}
Nobles, A.~L.; Dredze, M.; and Ayers, J.~W. 2019.
\newblock “Repeal and replace”: increased demand for intrauterine devices
  following the 2016 presidential election.
\newblock \emph{Contraception}.

\bibitem[{Nobles et~al.(2020)Nobles, Leas, Dredze, and
  Ayers}]{nobles2020examining}
Nobles, A.~L.; Leas, E.~C.; Dredze, M.; and Ayers, J.~W. 2020.
\newblock Examining Peer-to-Peer and Patient-Provider Interactions on a Social
  Media Community Facilitating Ask the Doctor Services.
\newblock In \emph{Proceedings of ICWSM}.

\bibitem[{Patel et~al.(2019)Patel, Blandford, Warner, Shawe, and
  Stephenson}]{patel2019feel}
Patel, D.; Blandford, A.; Warner, M.; Shawe, J.; and Stephenson, J. 2019.
\newblock ``I feel like only half a man'' Online Forums as a Resource for
  Finding a ``New Normal'' for Men Experiencing Fertility Issues.
\newblock In \emph{Proceedings of CSCW}.

\bibitem[{Pei and Jurgens(2020)}]{pei-jurgens-2020-quantifying}
Pei, J.; and Jurgens, D. 2020.
\newblock Quantifying Intimacy in Language.
\newblock In \emph{Proceedings of EMNLP}.

\bibitem[{Perrin and Anderson(2019)}]{perrin2019share}
Perrin, A.; and Anderson, M. 2019.
\newblock Share of US adults using social media, including Facebook, is mostly
  unchanged since 2018.
\newblock \emph{Pew Research Center}.

\bibitem[{Pfeffer et~al.(2022)Pfeffer, Mooseder, Hammer, Stritzel, and
  Garcia}]{Pfeffer2022ThisSS}
Pfeffer, J.; Mooseder, A.; Hammer, L.; Stritzel, O.; and Garcia, D. 2022.
\newblock This Sample seems to be good enough! Assessing Coverage and Temporal
  Reliability of Twitter's Academic API.
\newblock \emph{ArXiv}, abs/2204.02290.

\bibitem[{Pirolli and Card(2005)}]{pirolli2005sensemaking}
Pirolli, P.; and Card, S. 2005.
\newblock The sensemaking process and leverage points for analyst technology as
  identified through cognitive task analysis.
\newblock In \emph{Proceedings of International Conference on Intelligence
  Analysis}.

\bibitem[{Polis et~al.(2016)Polis, Bradley, Bankole, Onda, Croft, and
  Singh}]{polis2016typical}
Polis, C.~B.; Bradley, S.~E.; Bankole, A.; Onda, T.; Croft, T.; and Singh, S.
  2016.
\newblock Typical-use contraceptive failure rates in 43 countries with
  Demographic and Health Survey data: summary of a detailed report.
\newblock \emph{Contraception}.

\bibitem[{Rivas et~al.(2020)Rivas, Sadah, Guo, and
  Hristidis}]{rivas2020classification}
Rivas, R.; Sadah, S.~A.; Guo, Y.; and Hristidis, V. 2020.
\newblock Classification of Health-Related Social Media Posts: Evaluation of
  Post Content--Classifier Models and Analysis of User Demographics.
\newblock \emph{JMIR Public Health and Surveillance}.

\bibitem[{R\"{o}der, Both, and Hinneburg(2015)}]{roder2015exploring}
R\"{o}der, M.; Both, A.; and Hinneburg, A. 2015.
\newblock Exploring the Space of Topic Coherence Measures.
\newblock In \emph{Proceedings of WSDM}.

\bibitem[{Russo et~al.(2013)Russo, Parisi, Kukla, and Schwarz}]{russo2013women}
Russo, J.~A.; Parisi, S.~M.; Kukla, K.; and Schwarz, E.~B. 2013.
\newblock Women's information-seeking behavior after receiving contraceptive
  versus noncontraceptive prescriptions.
\newblock \emph{Contraception}.

\bibitem[{Samulowitz et~al.(2018)Samulowitz, Gremyr, Eriksson, and
  Hensing}]{Samulowitz2018BraveMA}
Samulowitz, A.; Gremyr, I.; Eriksson, E.~M.; and Hensing, G. 2018.
\newblock “Brave Men” and “Emotional Women”: A Theory-Guided Literature
  Review on Gender Bias in Health Care and Gendered Norms towards Patients with
  Chronic Pain.
\newblock \emph{Pain Research \& Management}.

\bibitem[{Sannon et~al.(2019)Sannon, Murnane, Bazarova, and
  Gay}]{sannon2019really}
Sannon, S.; Murnane, E.~L.; Bazarova, N.~N.; and Gay, G. 2019.
\newblock "I Was Really, Really Nervous Posting It": Communicating about
  Invisible Chronic Illnesses across Social Media Platforms.
\newblock In \emph{Proceedings of CHI}.

\bibitem[{Scarry(1987)}]{scarry1987body}
Scarry, E. 1987.
\newblock \emph{The body in pain: The making and unmaking of the world}.
\newblock Oxford University Press, USA.

\bibitem[{Schindler(2013)}]{schindler2013non}
Schindler, A.~E. 2013.
\newblock Non-contraceptive benefits of oral hormonal contraceptives.
\newblock \emph{International Journal of Endocrinology and Metabolism}.

\bibitem[{Schofield, Magnusson, and Mimno(2017)}]{schofield-etal-2017-pulling}
Schofield, A.; Magnusson, M.; and Mimno, D. 2017.
\newblock Pulling Out the Stops: Rethinking Stopword Removal for Topic Models.
\newblock In \emph{Proceedings of EACL}.

\bibitem[{Schofield and Mimno(2016)}]{schofield-mimno-2016-comparing}
Schofield, A.; and Mimno, D. 2016.
\newblock Comparing Apples to Apple: The Effects of Stemmers on Topic Models.
\newblock \emph{TACL}.

\bibitem[{Schofield, Thompson, and
  Mimno(2017)}]{schofield-etal-2017-quantifying}
Schofield, A.; Thompson, L.; and Mimno, D. 2017.
\newblock Quantifying the Effects of Text Duplication on Semantic Models.
\newblock In \emph{Proceedings of EMNLP}.

\bibitem[{Tangherlini(2000)}]{tangherlini2000heroes}
Tangherlini, T.~R. 2000.
\newblock Heroes and lies: Storytelling tactics among paramedics.
\newblock \emph{Folklore}.

\bibitem[{Weick, Sutcliffe, and Obstfeld(2005)}]{weick2005organizing}
Weick, K.~E.; Sutcliffe, K.~M.; and Obstfeld, D. 2005.
\newblock Organizing and the process of sensemaking.
\newblock \emph{Organization Science}.

\bibitem[{Yang et~al.(2019)Yang, Kraut, Smith, Mayfield, and
  Jurafsky}]{yang2019seekers}
Yang, D.; Kraut, R.~E.; Smith, T.; Mayfield, E.; and Jurafsky, D. 2019.
\newblock Seekers, providers, welcomers, and storytellers: Modeling social
  roles in online health communities.
\newblock In \emph{Proceedings of CHI}.

\bibitem[{Yee and Simon(2010)}]{yee2010role}
Yee, L.; and Simon, M. 2010.
\newblock The role of the social network in contraceptive decision-making among
  young, African American and Latina women.
\newblock \emph{Journal of Adolescent Health}.

\bibitem[{Yoost(2014)}]{yoost2014understanding}
Yoost, J. 2014.
\newblock Understanding benefits and addressing misperceptions and barriers to
  intrauterine device access among populations in the United States.
\newblock \emph{Patient Preference and Adherence}.

\bibitem[{Young and Miller(2019)}]{young2019girl}
Young, A.~L.; and Miller, A.~D. 2019.
\newblock "This Girl is on Fire": Sensemaking in an Online Health Community for
  Vulvodynia.
\newblock In \emph{Proceedings of CHI}.

\bibitem[{Yun(2019)}]{yun2019decline}
Yun, H. 2019.
\newblock The Decline of WebMD?
\newblock \emph{NYC Data Science Academy: Data Science Blog}.

\bibitem[{Zhang, N.~Bazarova, and Reddy(2021)}]{zhang2021distress}
Zhang, R.; N.~Bazarova, N.; and Reddy, M. 2021.
\newblock Distress Disclosure across Social Media Platforms during the COVID-19
  Pandemic: Untangling the Effects of Platforms, Affordances, and Audiences.
\newblock In \emph{Proceedings of CHI}.

\end{thebibliography}
